\documentclass[letterpaper, 10 pt, conference]{ieeeconf}  

\IEEEoverridecommandlockouts                              

\usepackage{graphicx}
\usepackage{amsmath,amsfonts,amssymb}
\usepackage{multirow}
\usepackage[dvipsnames]{xcolor}
\usepackage{tabularx}
\usepackage{tikz,tkz-euclide}
\usepackage{caption}
\tikzset{>=latex}
\usepackage{xcolor}
\usepackage[hidelinks]{hyperref}

\title{\LARGE \bf Towards Robust Perception for Assistive Robotics: An RGB-Event-LiDAR Dataset and Multi-Modal Detection Pipeline}

 \author{Adam Scicluna, Cedric Le Gentil, Sheila Sutjipto and Gavin Paul
\thanks{All authors are with the Robotics Institute, Faculty of Engineering and Information Technology, University of Technology Sydney (UTS), Australia. Corresponding author: {\tt\footnotesize \{adam.scicluna@alumni.uts.edu.au\}}}
\thanks{Cedric Le Gentil is supported by the Australian Research Council Discovery Project under Grant DP210101336. Sheila Sutjipto is supported by Australian Government Research Training Program Scholarships. The authors would like to acknowledge the support from the ARC Industrial Transformation Training Centre (ITTC) for Collaborative Robotics in Advanced Manufacturing under grant IC200100001.}
\thanks{© 2024 IEEE.  Personal use of this material is permitted.  Permission from IEEE must be obtained for all other uses, in any current or future media, including reprinting/republishing this material for advertising or promotional purposes, creating new collective works, for resale or redistribution to servers or lists, or reuse of any copyrighted component of this work in other works.}
}

\begin{document}

\maketitle
\thispagestyle{empty} 
\pagestyle{empty} 

\begin{abstract}

The increasing adoption of human-robot interaction presents opportunities for technology to positively impact lives, particularly those with visual impairments, through applications such as guide-dog-like assistive robotics. We present a pipeline exploring the perception and ``intelligent disobedience" required by such a system. A dataset of two people moving in and out of view has been prepared to compare RGB-based and event-based multi-modal dynamic object detection using LiDAR data for 3D position localisation. Our analysis highlights challenges in accurate 3D localisation using 2D image-LiDAR fusion, indicating the need for further refinement. Compared to the performance of the frame-based detection algorithm utilised (YOLOv4), current cutting-edge event-based detection models appear limited to contextual scenarios, such as for automotive platforms. This is highlighted by weak precision and recall over varying confidence and Intersection over Union (IoU) thresholds when using frame-based detections as a ground truth. Therefore, we have publicly released this dataset to the community, containing RGB, event, point cloud and Inertial Measurement Unit (IMU) data along with ground truth poses for the two people in the scene to fill a gap in the current landscape of publicly available datasets and provide a means to assist in the development of safer and more robust algorithms in the future: \url{https://uts-ri.github.io/revel/}.

\end{abstract}

\section{Introduction}

Training guide dogs requires significant time and financial resources \cite{GuideDogsVictoria2023}, with only about half of the dogs successfully completing the programs \cite{Tomkins2012-GuideDogSuccess}. This creates a gap between the availability of guide dogs and the needs of visually impaired individuals. Thus, there is growing interest in cost-effective robotic alternatives \cite{Hong2022-SystematicLitReview}. These robotic substitutes must replicate the ``intelligent disobedience" of guide dogs, where the robot refuses unsafe commands based on its understanding of an environment. Therefore, reliable perception and decision-making algorithms are necessary to ensure user safety.

\begin{figure}
    \centering
    \def\vdist{0.5cm}
    \def\hdist{0.1cm}
    \def\legenddist{-0.1cm}
    \def\legendsize{\scriptsize}
    \def\height{5cm}
    \def\width{\columnwidth}
    \begin{tikzpicture}
        
        \tikzstyle{legend} = [draw=none, minimum height = 0.2em, text width = 8em,  minimum width = 5.5em, align =center, anchor=center, execute at begin node=\setlength{\baselineskip}{8pt}]
        
        \node[anchor=south west, inner sep=0, outer sep=0] (image1) { \includegraphics[width=\width]{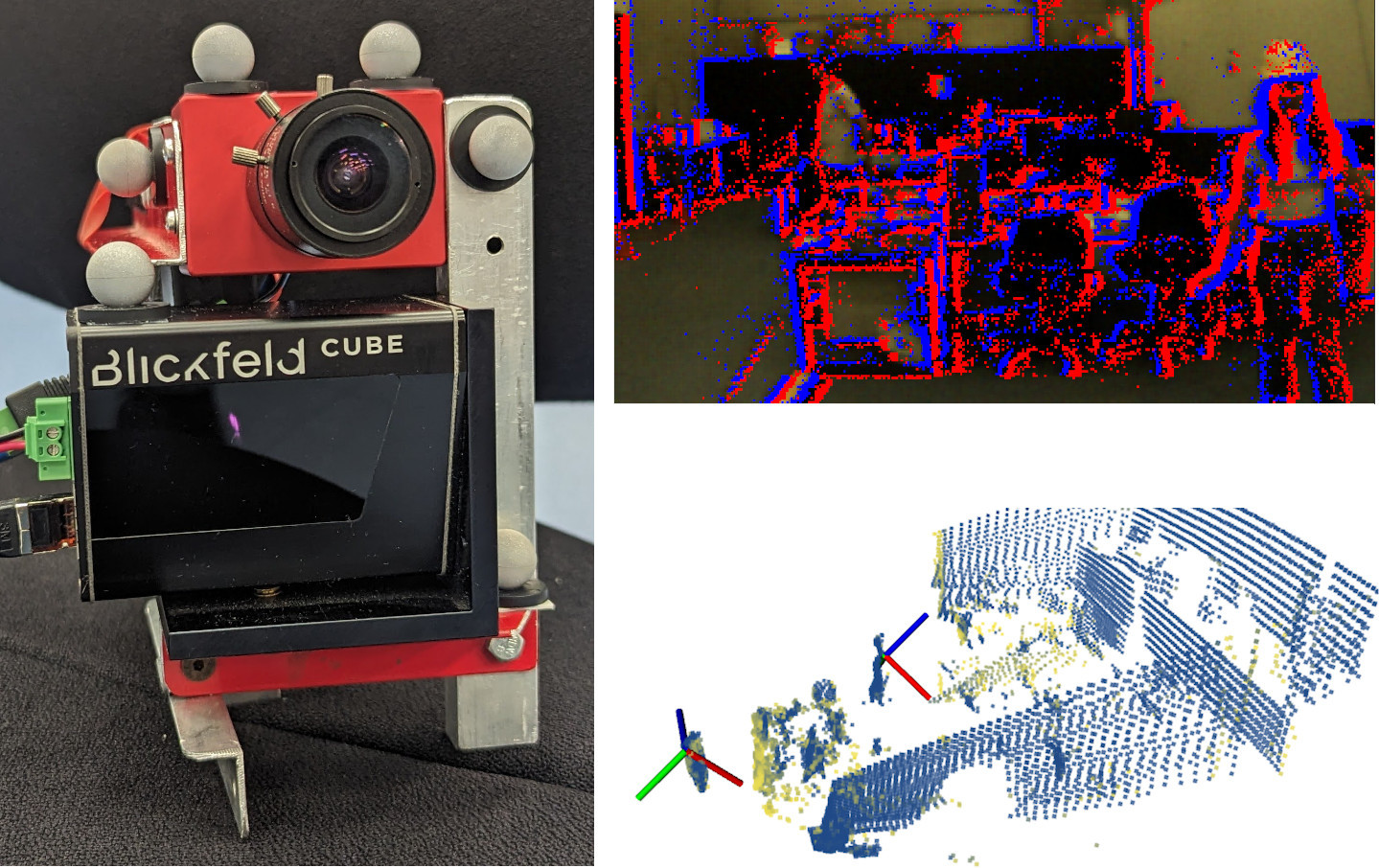}};
        \node[legend] at (0.21*\width, -0.14) {\legendsize(a) Sensor suite} ;

        \node[legend] at (0.72*\width, 2.76) {\legendsize (b) Vision data sample};
        \node[legend] at (0.72*\width, -0.14){\legendsize (c) Geometric data sample};
    \end{tikzpicture}
    \vspace{-0.5cm}
    \caption{(a) Sensor suite featuring a DAVIS 346 frame-event camera and a Cube1 LiDAR for dataset collection. (b) DAVIS camera data sample: events (polarity-coloured in red or blue) overlaid on the RGB frame. (c) LiDAR scan sample with object motion-captured ground truth poses (frames).}
    \label{fig:teaser}
\end{figure}
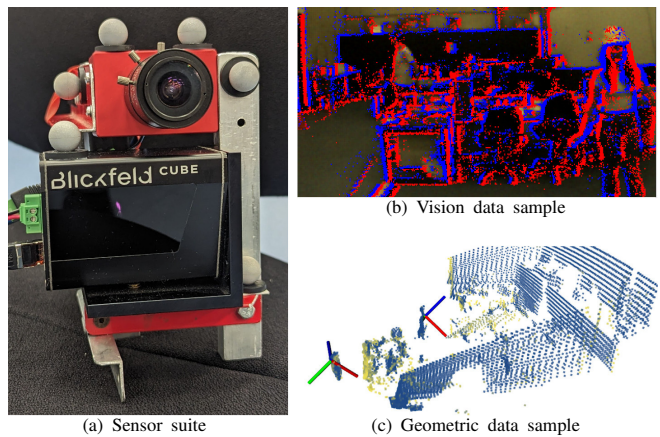

Perception and scene understanding are essential capabilities for robotic systems to be integrated into daily life.
Such systems must robustly comprehend their surroundings in real-time for subsequent decision-making and actions toward safe operations.
In the context of guide-dog-like assistive robotics, this translates into the accurate detection and identification of both static and dynamic objects such as cars, bicycles, pedestrians, etc, and the ability to estimate their 3D pose and dynamics. This would enable the robot to navigate and avoid hazards, enhancing safety and usability.

Recent machine learning advances and GPU availability have enabled efficient context and scene recognition from RGB images using neural networks~\cite{Krizhevsky2012ImageNet,Ren2017FasterR}.
Yet, controlling movement or manipulation demands 3D pose knowledge. Although RGBD and stereo cameras provide accurate depth information, their low dynamic range and susceptibility to motion blur, along with RGBD cameras' poor outdoor performance due to sunlight's infrared radiation, pose challenges in safety-critical, dynamic or brightly lit environments \cite{Paul2016}. To address this, LiDAR-camera multi-modal sensor suites are used, though LiDAR's sparsity, noise, and slow acquisition rate complicate detection.
Thus, RGB-LiDAR systems still struggle with standard cameras' limitations in scenarios under high dynamic range or low light conditions.

Event-based cameras~\cite{Lichtsteiner2008DVS} offer a solution to the limitations of traditional frame-based vision by capturing pixel-level changes in illumination independently. However, object detection methods for event cameras are still in their early stages compared to those for RGB data~\cite{Redmon2016YouOnly, Ren2017FasterR}. To fully exploit the benefits of event cameras, developing new algorithms tailored to their unique data output is crucial.

Early RGB-based object detection used handcrafted features and classic machine learning but struggled with variability and extensive parameter tuning. CNNs revolutionised the field with superior performance in managing data variation~\cite{Krizhevsky2012ImageNet}, greatly aided by extensively labelled public datasets~\cite{Deng2009ImageNet, Lin2014Microsoft, Kuznetsova2020Open, Everingham2010Pascal}, where millions of images across thousands of categories provide a crucial benchmark for performance evaluation.  Tailored datasets like KITTI~\cite{Geiger2013Vision} for autonomous driving have also been pivotal in advancing algorithm development. Prominent approaches include R-CNN and its iterations~\cite{Girshick2015FastRCnn, Ren2017FasterR}, where selective searches generate regions classified by a CNN. Recent efforts have focused on improving efficiency evident with YOLO \cite{Bochkovskiy2020-YOLOv4} and single shot detectors~\cite{Liu2016SSD, Wang2018Pelee}.
Works have also combined object detection with semantic segmentation to enhance scene awareness and object analysis~\cite{Dai2016Instance, He2017MaskRCNN}.

Frame-based algorithms do not translate well for use with event camera data. Consequently, efforts have been made to reconstruct dense greyscale images from sparse event data to feed to a CNN, introducing an intermediate step that increases the computational burden and latency. Alternatively, CNNs~\cite{Gehrig2019EndToEnd}, spiking neural networks~\cite{Gehrig2020VelocitySpiking}, and graph neural networks~\cite{Schaefer2022Asynchronous} have demonstrated object detection in the event space. Recently, transformers like RVTs have achieved state-of-the-art performance on the Gen1 and 1-Mpx~\cite{Perot2020Learning} datasets~\cite{De2020LargeGen1, Perot2020Learning, Gehrig2023-RVT}. While promising, these methods lack the accessibility of algorithms like YOLO~\cite{Redmon2016YouOnly}. 
Furthermore, event camera-based techniques struggle to generalise to other scenes due to the limited variety in existing datasets.

Leveraging the complementary nature of LiDAR point clouds and RGB images enables a more comprehensive scene understanding~\cite{Zhao2023LIFSeg}. For object detection, strategies include using RGB-trained networks on image-like data generated from LiDAR data \cite{dai2018connecting}, constructing pseudo-LiDAR data from RGB images~\cite{Lin2022CL3D}, using 2D detectors to propose 3D search spaces~\cite{Qi2018FPNet}, and employing RVTs~\cite{Bai2022TransFusion}. Fusing LiDAR and camera data has proven effective for semantic segmentation, using methods like mapping LiDAR points to the output of an image-based semantic segmentation network and inputting the data into a LiDAR detector~\cite{Vora2020PointPainting}, and addressing sparsity with cylindrical partitioning and asymmetrical 3D CNNs~\cite{Zhu2022Cylindrical}.

To our knowledge, no publicly available dataset contains data from an event camera, an RGB camera, a LiDAR, and an IMU, while providing ground truth poses of the sensor suite and dynamic objects.
In this paper, we introduce a labelled dataset and propose a multi-modal perception pipeline for 3D object detection and spatial pose estimation that combines a 2D detection step (based on RGB or event vision) and a depth estimation step using LiDAR data.
Fig.~\ref{fig:teaser} shows our sensor and some data samples.
We evaluate the performance using the event and RGB camera, highlighting the potential and challenges for future robotic guide-dog systems.

\section{Dataset}

\subsection{Sensor suite and data collection}

The dataset introduced in this paper is collected indoors with a handheld sensor suite moving in the field of view of a \textit{Vicon} motion-capture system.
Two people, also tracked by the motion-capture system, are moving in and out of the sensor suite field of view.
The sensor suite consists of:
\begin{itemize}
    \item \textbf{Inivation DAVIS346} event camera:
        \textit{Stream of event} data (up to 1MHz) with each event being a tuple of $x$ and $y$ positions in the image space, $t$ the timestamp, and $p$ the polarity of the corresponding illumination change; 
        \textit{RGB images} at 23Hz;
        \textit{6-DoF IMU} at 1kHz (3-axis gyroscope and 3-axis accelerometer).
    \item \textbf{Blickfeld Cube1} LiDAR: \textit{3D point clouds} at 7.9Hz with point-wise timestamps.
\end{itemize}
All the sensor's measurements and the output of the motion-capture system are recorded with ROS.
We use the \textit{rpg\_dvs\_ros} driver for the DVS camera and the Blickfeld ROS driver for the LiDAR.
As illustrated in Fig.~\ref{fig:teaser}, the sensor suite is equipped with a set of reflective markers tracked by the \textit{Vicon} system.
Similarly, the people moving in the surroundings wear helmets with reflective markers.
Subsequently, the \textit{Vicon} system provides the 6-DoF pose of the 2 persons and sensor suite in an arbitrarily fixed reference frame.
Overall, the dataset spans 14 minutes over four ROSBags, containing approximately 774 million events\footnote{The DVS driver cuts the event stream into variable-length event-array messages; thus, the dataset contains 25000 event-array messages.}, 22000 RGB images, 6700 point clouds, and 70000 ground truth poses each for two persons in the scene. For the experimentation performed, the ROSBag entitled ``dynamic.bag" was used. For convenience and utility, the dataset is labelled with the class identifier corresponding to the colour helmet worn by the person.

\subsection{Calibration}
\label{sec:calibration}

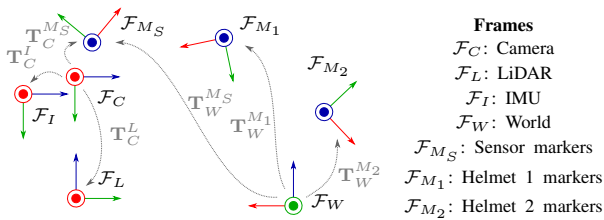
\begin{figure}
    \centering
    \def\textsize{\scriptsize}
    \def\legendsize{\scriptsize}
    \def\scale{0.6}
    \def\legendx{4.0}
    \def\legendy{1.15}
    \def\vdist{-0.12}
    \begin{tikzpicture}[auto]
        \tikzstyle{legend} = [draw=none, rectangle, minimum height = 0.5em, text width = 9.5em,  minimum width = 9.5em, align = center, node distance = 5em, execute at begin node=\setlength{\baselineskip}{8pt}]
    
        \node{\includegraphics[clip, width=\scale\columnwidth]{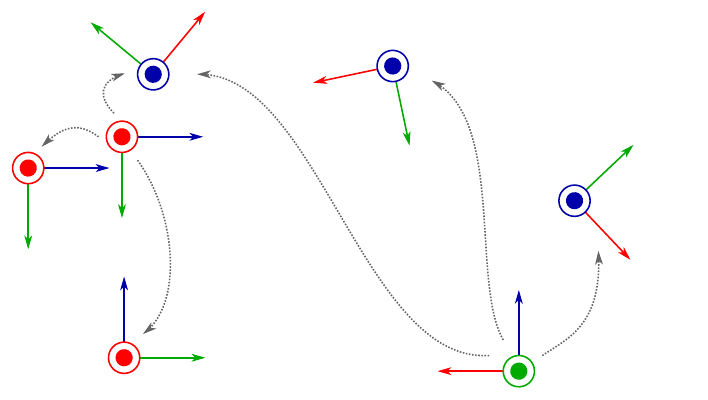}};
        \node[draw=none] at (-2.0*\scale,0.3*\scale) {\textsize$\mathcal{F}_C$};
        \node[draw=none] at (-2.0*\scale,-1.5*\scale) {\textsize$\mathcal{F}_L$};
        \node[draw=none] at (-3.5*\scale,-0.1*\scale) {\textsize$\mathcal{F}_I$};
        \node[draw=none] at (-1.3*\scale,1.9*\scale) {\textsize$\mathcal{F}_{M_S}$};
        \node[draw=none] at (1.3*\scale,1.9*\scale) {\textsize$\mathcal{F}_{M_{1}}$};
        \node[draw=none] at (2.8*\scale,1.0*\scale) {\textsize$\mathcal{F}_{M_{2}}$};
        \node[draw=none] at (2.8*\scale,-2.0*\scale) {\textsize$\mathcal{F}_{W}$};
        \node[draw=none] at (-4.0*\scale,1.2*\scale) {\color{gray}\textsize$\mathbf{T}_C^I$};
        \node[draw=none] at (-1.7*\scale,-0.5*\scale) {\color{gray}\textsize$\mathbf{T}_C^L$};
        \node[draw=none] at (-3.4*\scale,1.7*\scale) {\color{gray}\textsize$\mathbf{T}_C^{M_S}$};
        \node[draw=none] at (0.2*\scale,0.2*\scale) {\color{gray}\textsize$\mathbf{T}_W^{M_S}$};
        \node[draw=none] at (1.1*\scale,-0.3*\scale) {\color{gray}\textsize$\mathbf{T}_W^{M_{1}}$};
        \node[draw=none] at (3.5*\scale,-1.4*\scale) {\color{gray}\textsize$\mathbf{T}_W^{M_{2}}$};

        \node[legend] at (\legendx, \legendy) (title) {\legendsize \textbf{Frames}};
        \node[legend, below=\vdist of title] (cam) {\legendsize $\mathcal{F}_C$: Camera};
        \node[legend, below=\vdist of cam] (lidar) {\legendsize $\mathcal{F}_L$: LiDAR};
        \node[legend, below=\vdist of lidar] (imu) {\legendsize $\mathcal{F}_I$: IMU};
        \node[legend, below=\vdist of imu] (world) {\legendsize $\mathcal{F}_{W}$: World};
        \node[legend, below=\vdist of world] (smarker) {\legendsize $\mathcal{F}_{M_S}$: Sensor markers};
        \node[legend, below=\vdist of smarker] (ymarker) {\legendsize $\mathcal{F}_{M_1}$: Helmet 1 markers};
        \node[legend, below=\vdist of ymarker] (gmarker) {\legendsize $\mathcal{F}_{M_2}$: Helmet 2 markers};
        
    \end{tikzpicture}
    \vspace{-0.2cm}
    \caption{Frames and geometric transformations in the dataset.}
    \label{fig:calibration}
\end{figure}

To use our dataset effectively, we must first perform the intrinsic calibration of the camera and extrinsic calibration between the various sensors and the set of reflective markers.
Fig.~\ref{fig:calibration} shows the set of geometric transformations $\mathbf{T}_a^b$ estimated during  calibration ($\mathbf{T}_C^{L}$, $\mathbf{T}_C^{M_S}$, and $\mathbf{T}_C^{I}$) or given by the Vicon system ($\mathbf{T}_W^{M_S}$, $\mathbf{T}_W^{M_1}$, and $\mathbf{T}_W^{M_2}$).
Note that as the RGB and event data are being collected by the same cells in the \textit{DAVIS346}, the reference frame of the event and RGB camera are collocated (labelled ``Camera" in Fig.~\ref{fig:calibration}).
Thus, the intrinsic calibration parameters obtained with the RGB camera apply to the event camera as both data types share the same optical path.
Calibration sequences and parameter estimates are included with the main dataset.
Table~\ref{table:calibration} details the estimation process for each transformation \footnote{The camera's intrinsics are obtained with Matlab's calibration toolbox https://au.mathworks.com/help/vision/ref/cameracalibrator-app.html}.

\begin{table}
\caption{Insight into the extrinsic calibration procedure of the sensor suite used to collect the proposed dataset.}
\def\textsize{\scriptsize}
\centering
\renewcommand\tabularxcolumn[1]{m{#1}}
\newcolumntype{Y}{>{\centering\arraybackslash}X}
\newcolumntype{C}[1]{ >{\centering\arraybackslash} m{#1}}
\begin{tabularx}{1\linewidth}{C{1.5cm} Y }
\hline
\textbf{Trans.} &
  \textbf{Details} 
\\ \hline
    $\mathbf{T}_C^{M_S}$ & {\textsize Camera position from checkerboard detection, then eye-in-hand calibration with Vicon poses of $\mathcal{F}_{M_S}$}
\\ \hline
    $\mathbf{T}_C^L$ & {\textsize Checkerboard plane equation from camera and point-to-plane minimisation with LiDAR points on the checkerboard}
\\ \hline
    $\mathbf{T}_C^I$ & {\textsize Kalibr\footnote{https://github.com/ethz-asl/kalibr}: Checkerboard for camera position and continuous-time batch state estimation}
\\ \hline
\end{tabularx}

\label{table:calibration}
\end{table}

\section{Multi-modal scene understanding}

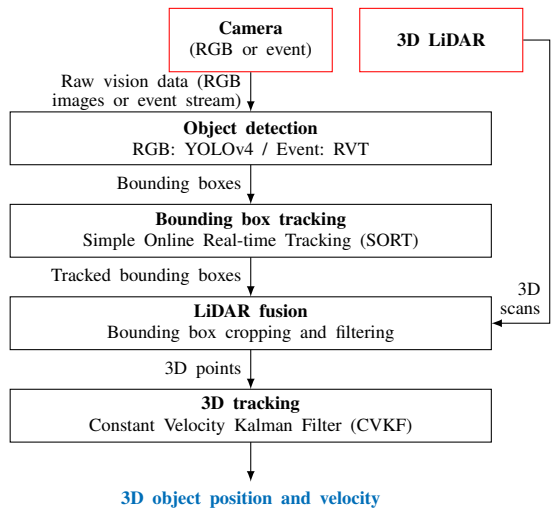
\begin{figure}
    \centering
    \def\hdist{1em}
    \def\vdist{1.5em}
    \def\blockheight{2em}
    \def\blockwidth{18em}
    \def\innerpad{0.1em}
    \def\textsize{\scriptsize}
    \begin{tikzpicture}[auto]
    \tikzstyle{input} = [draw, fill=white, rectangle, minimum height = 2.4em, text width = 5.5em,  minimum width = 5.5em, align = center, node distance = 5em, draw=red, execute at begin node=\setlength{\baselineskip}{8pt}]
    \tikzstyle{block} = [draw, fill=white, rectangle, minimum height = \blockheight, text width = \blockwidth,  minimum width = \blockwidth, align = center, inner sep=\innerpad, outer sep=0, node distance = 0em, execute at begin node=\setlength{\baselineskip}{8pt}] 
    \tikzstyle{output} = [draw=none, fill=white, text=NavyBlue, rectangle, minimum height = 1em, text width =\blockwidth,  minimum width =\blockwidth, align = center, node distance = 11em, execute at begin node=\setlength{\baselineskip}{8pt}] 

    \node [input] (cam) {\textsize \textbf{Camera} \\(RGB or event)};
    \node [block, below = \vdist of cam] (detection) {\textsize \textbf{Object detection}\\RGB: YOLOv4 / Event: RVT};
    \node [block, below=\vdist of detection] (sort) {\textsize \textbf{Bounding box tracking}\\Simple Online Real-time Tracking (SORT)};
    \node [block, below= \vdist of sort] (fusion) {\textsize \textbf{LiDAR fusion}\\Bounding box cropping and filtering};
    \node [block, below= \vdist of fusion] (tracking) {\textsize \textbf{3D tracking}\\Constant Velocity Kalman Filter (CVKF)};
    \node [input, right=\hdist of cam] (lidar) {\textsize \textbf{3D LiDAR}};
    \node [output, below=\vdist of tracking] (result) {\textsize \textbf{3D object position and velocity}};

    \draw[->] (cam) -- node[left, align=right, text width = 9em, execute at begin node=\setlength{\baselineskip}{7pt}]{\textsize Raw vision data (RGB images or event stream)} (detection);
    \draw[->] (detection) -- node[left, align=right, text width = 9em, execute at begin node=\setlength{\baselineskip}{7pt}]{\textsize Bounding boxes} (sort);
    \draw[->] (sort) -- node[left, align=right, text width = 9em, execute at begin node=\setlength{\baselineskip}{7pt}]{\textsize Tracked bounding boxes} (fusion);
    \draw[->] (fusion) -- node[left, align=right, text width = 9em, execute at begin node=\setlength{\baselineskip}{7pt}]{\textsize 3D points} (tracking);

    \coordinate[right=1em of lidar] (pathanchor);
    \draw[->] (lidar.east) -- (pathanchor) |- node[above left, align=right, text width = 2em, execute at begin node=\setlength{\baselineskip}{7pt}]{\textsize 3D scans} (fusion.east);

    \draw[->] (tracking) -- (result);
    
    \end{tikzpicture}
    \vspace{-0.2cm}
    \caption{Block diagram overview of the proposed vision-LiDAR object detection and tracking.}
    \label{fig:overview}
\end{figure}

To enable downstream applications such as guide-dog-like assistive robots, we explore fusing vision and LiDAR data for the 3D localisation of pedestrians and vehicles. The methodology can also be applied to static objects such as trees, buildings and roads. While a complete system should include these necessities, the motivation of this work is to focus on dynamic objects due to the further requirement of tracking relative motion.
Fig.~\ref{fig:overview} presents an overview of the proposed pipeline.
The main steps are, first, the vision-based detection of objects in the image space (2D), followed by the tracking of the resulting bounding boxes with the Simple Online and Real-time Tracking (SORT) algorithm~\cite{Bewley2016-SORT}.
Then, the bounding boxes are used to crop and filter the LiDAR scans before performing state estimation in the 3D space using a Constant-Velocity Kalman Filter (CVKF).
Note that this pipeline can be used with an event camera or a standard RGB camera if the detection algorithm provides bounding boxes around the detected objects.
The rest of this section provides details about the components of the proposed pipeline. 

\subsection{2D object detection and tracking}

\subsubsection{RGB-based detection}

When the proposed framework is used with an RGB camera, we use YOLOv4 \cite{Bochkovskiy2020-YOLOv4} for the task of object detection.
YOLOv4 is a CNN-based algorithm renowned for its real-time capabilities and accuracy.
It is trained on the MS-COCO dataset~\cite{Lin2014Microsoft} and can classify 80 different types of objects, including pedestrians and vehicles.
Its one-shot detection approach surpasses traditional two-shot detectors like Faster R-CNN in terms of inference speed.
The RGB images from the DAVIS346 are undistorted using the camera's intrinsic parameters before being passed to YOLOv4.
The output consists of 2D bounding boxes.

\subsubsection{Event-based detection}

For event-based vision, our pipeline relies on RVT~\cite{Gehrig2023-RVT}.
The choice of RVT is motivated by its proficiency in detecting both vehicle and pedestrian data, coupled with its fast inference time relative to alternative event-based models.
RVT relies on recurrent transformers to leverage the spatiotemporal nature of event data.
Accordingly, the stream of events is preprocessed into a succession of 4-dimensional tensors of size ($2, T, h, w$), with $h$ and $w$ the resolution of the camera, by binning the events into $T$ temporal slices (10 slices within a 50 ms window in the publicly available model). 
The first dimension of the tensor represents the two polarities of the events, thus storing the events triggered by positive and negative changes separately.
The authors of RVT have released pre-trained models \textit{Gen1} and \textit{1-Mpx} that are trained with the Gen1~\cite{De2020LargeGen1} and 1-Mpx~\cite{Perot2020Learning} automotive datasets, respectively.
To infer objects' bounding boxes using the DAVIS346 data, we crop or pad the event tensors to fit the required input size of the models ($h,w$).

\subsubsection{2D tracking}
\label{sec:2d_tracking}

The 2D bounding boxes in the image from the event and RGB object detectors are quite noisy in regards to the position and amount of misdetection.
The proposed pipeline leverages the SORT algorithm for multi-frame object association and position estimation to address this issue.
SORT employs the Hungarian Method in conjunction with a Linear CVKF to track objects across frames independently of other objects and camera motion.
The state vector in the CVKF is $x = [u, v, s, r, \dot{u}, \dot{v}, \dot{s}]^\top$,
where $u$, $v$, $\dot{u}$ and $\dot{v}$ represent the centre coordinates and velocity in pixels/frame of the bounding box, $s$ and $\dot{s}$ represent the bounding box area and change in area respectively, and $r$ represents the aspect ratio of the bounding box, which is assumed to be constant. We apply small changes to parameters outlined in Table~\ref{table:sort_parameters} to better handle occlusions and instances of missed subsequent associations to a tracker, which are dangers for a guide-dog-like aid.

\subsection{3D fusion}

\subsubsection{LiDAR scan filtering}

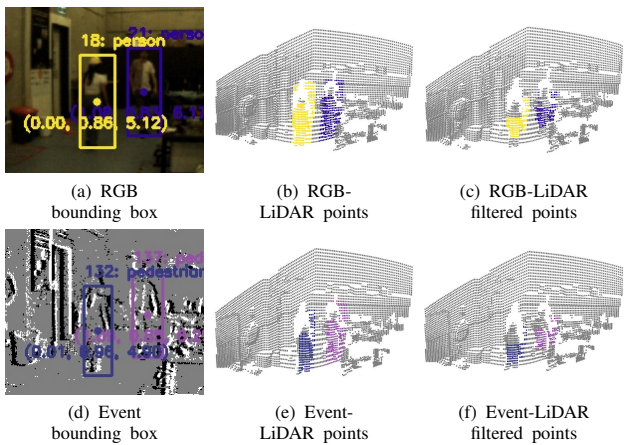
\begin{figure}
    \centering
    \def\vdist{0.5cm}
    \def\hdist{-0.1cm}
    \def\legenddist{-0.1cm}
    \def\legendsize{\scriptsize}
    \def\height{2.2cm}
    \begin{tikzpicture}
        \tikzstyle{legend} = [draw=none, minimum height = 0.2em, text width = 6.5em,  minimum width = 6.5em, align =center, execute at begin node=\setlength{\baselineskip}{8pt}]
    
        \node[] (rgbimg) {\includegraphics[clip, trim=0cm 0cm 0cm 0cm, height=\height]{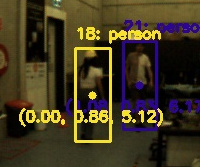}};
        \node[right=\hdist of rgbimg] (rgbbb) {\includegraphics[clip, trim=0cm 0cm 0cm 0cm, height=\height]{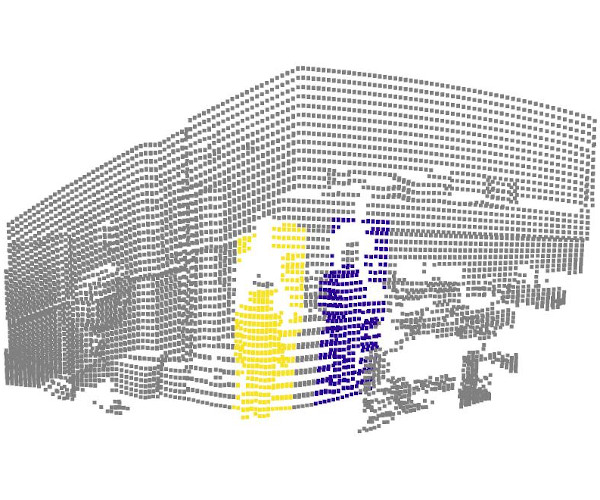}};
        \node[right=\hdist of rgbbb] (rgbcs) {\includegraphics[clip, trim=0cm 0cm 0cm 0cm, height=\height]{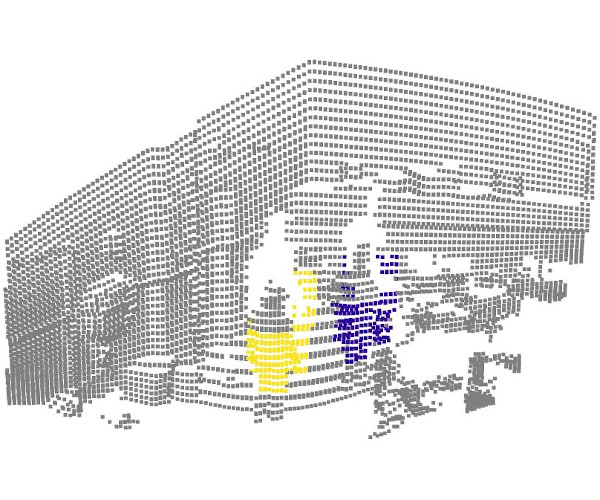}};
        \node[below=\vdist of rgbimg] (eventimg) {\includegraphics[clip, trim=0cm 0cm 0cm 0cm, height=\height]{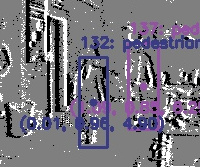}};
        \node[right=\hdist of eventimg] (eventbb) {\includegraphics[clip, trim=0cm 0cm 0cm 0cm, height=\height]{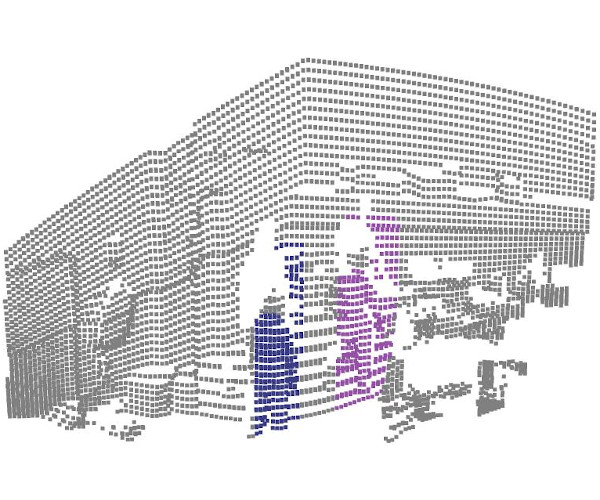}};
        \node[right=\hdist of eventbb] (eventcs) {\includegraphics[clip, trim=0cm 0cm 0cm 0cm, height=\height]{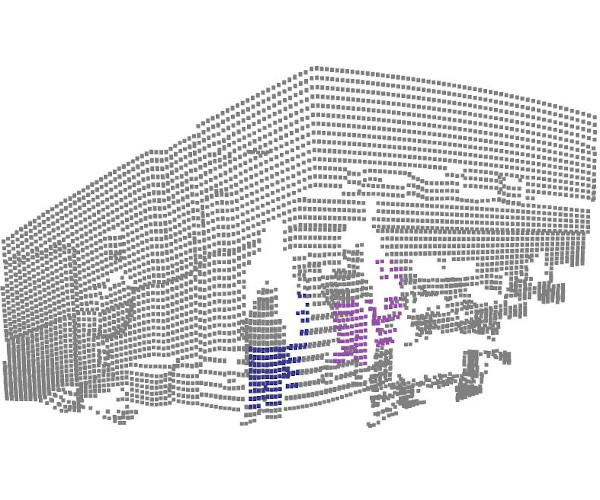}};

        \node[legend, below=\legenddist of rgbimg]{\legendsize (a) RGB bounding box};
        \node[legend, below=\legenddist of rgbbb]{\legendsize (b) RGB-LiDAR points};
        \node[legend, below=\legenddist of rgbcs]{\legendsize (c) RGB-LiDAR filtered points};
        \node[legend, below=\legenddist of eventimg]{\legendsize (d) Event bounding box};
        \node[legend, below=\legenddist of eventbb]{\legendsize (e) Event-LiDAR points};
        \node[legend, below=\legenddist of eventcs]{\legendsize (f) Event-LiDAR filtered points};
        
    \end{tikzpicture}
    \vspace{-0.2cm}
    \caption{Bounding box point cloud segmentation and filtering examples via vision-based (RGB and event) object detection.}
    \label{fig:bbox_pc}
\end{figure}

Given stable 2D bounding boxes from the aforementioned vision-based detector-and-tracking step, we first select the LiDAR points that fall into a bounding box by projecting each point of a LiDAR scan into the image using $\mathbf{T}_C^L$ and the camera intrinsics as illustrated in Fig.~\ref{fig:bbox_pc}
Unfortunately, the points associated with a bounding box do not only correspond to the detected object but also to the foreground and background.
Accordingly, we propose a simple filtering method to only extract points belonging to the detected object.
Based on the assumption that the centre of the detected object is roughly aligned with the centre of the bounding box, only points present in a square around the bounding box centre are considered.
The ratio of the square's area to the bounding box's area is scaled linearly with the ratio of the bounding box's area to the image resolution. Therefore, as the bounding box gets smaller, the ratio of the square to bounding box area increases, and vice-versa.
For the rest of the pipeline, the object position is represented with a single 3D point.
Accordingly, we use the median of the points inside the square to feed the tracker presented in the following subsection.

\subsubsection{3D tracking}

Provided with the point representation from the LiDAR scan filtering and the bounding box tracking ID from Section~\ref{sec:2d_tracking}, the proposed pipeline initialises and maintains independent CVKF for each object track.
Inspired by the work in \cite{Weng2020-AB3DMOT},
the CVKF state vector consists of the object position and velocity: $x_{3D} = [x, y, z, \dot{x}, \dot{y}, \dot{z}]^\top$.

\section{Experiments}

\subsection{Implementation}
The quantitative results are obtained using the proposed dataset collected with an \textit{Inivation DAVIS346} camera and a \textit{Blickfeld Cube1} LiDAR.
For the event-based detector, we empirically chose the 1-Mpx model of RVT \cite{Gehrig2023-RVT} after testing Gen1 and 1-Mpx using our dataset.
No significant performance difference was found.
Both RGB-based and event-based object detectors were used without any retraining or fine-tuning of the networks' weights.
Table~\ref{table:sort_parameters} shows the parameters of the SORT algorithm for vision-based tracking (Section~\ref{sec:2d_tracking}).
The RGB and event detectors have different noise characteristics, so SORT tracker parameters differ slightly.
\begin{table}[hb]
\caption{The SORT algorithm parameters for image space object tracking.}
\centering
\newcolumntype{Y}{>{\centering\arraybackslash}X}
\newcolumntype{C}[1]{>{\centering\let\newline\\\arraybackslash\hspace{0pt}}m{#1}}
\begin{tabularx}{1\linewidth}{C{6cm} Y Y }
\hline
\textbf{Parameter} &
  \textbf{RGB} &
  \textbf{Event} 
\\ \hline
    Maximum age of unmatched tracker [no. of frames] &
    10 &
    10 
\\ \hline
    Maximum unmatched predictions [no. frames]  &
    5 &
    3
\\ \hline
    Min. number of associated detections for tracking &
    3 &
    1
\\ \hline
  Min. number of previous associations for prediction &
  10 &
  1
\\ \hline
  IoU threshold for association &
  0.3 &
  0.3
\\ \hline
\end{tabularx}
\label{table:sort_parameters}
\end{table}
Experiments were conducted on a low-performance laptop with Ubuntu 20.04.6 LTS, an NVIDIA GTX GeForce 1650 GPU, an AMD Ryzen 7 5700U CPU, and 16GB of RAM.
The proposed pipeline runs close to real-time with both RGB-based and event-based detection.

\subsection{Vision-based object detection}
To evaluate the performance of the vision-based object detectors in our pipeline for assistive robotics in dynamic settings, we performed object detection with both YOLOv4 and RVT using the proposed dataset. We only consider YOLOv4 detections above a confidence score of 0.5 while varying the RVT confidence threshold across evaluations.

Each RGB frame is associated with a 50~ms event tensor/sequence required for RVT's prediction.
Table~\ref{table:change_iou} displays RVT's precision and recall (confidence threshold of 0.3), with and without the tracking, using  YOLOv4 as the ground truth due to its proven accuracy~\cite{Bochkovskiy2020-YOLOv4}.
The definitions of true/false positive/negative for precision and recall are based on IoU thresholds between the bounding boxes of both methods.
Table~\ref{table:change_iou} displays results for varying IoU thresholds, while Table~\ref{table:change_conf} shows results for a fixed IoU threshold with varying RVT confidence thresholds.

The results show that YOLOv4 outperforms the RVT model. The precision scores suggest a higher number of misclassifications with erroneous class attribution.
Interestingly, using SORT increased the recall but decreased the precision. This suggests that when true positive detections occur in one sequence of events but not in the ensuing sequences, the SORT algorithm improves the detection rate due to its ability to predict the subsequent positions of an object, reducing the number of false negatives.
However, when the tracker incorrectly estimates the object dynamics or false positive detections occur, the precision score decreases.

\begin{table*}
\caption{Evaluation of event-based detection vs. YOLOv4: varying IoU thresholds @ confidence threshold = 0.3.}
\centering
\begin{tabular}{ccccccccccc}
\multicolumn{2}{l}{\multirow{2}{*}{}}                          & \multicolumn{9}{l}{\textbf{Event-Based Detection: IoU thresholds @ Confidence threshold = 0.3}} \\
\multicolumn{2}{l}{} & \textbf{0.5} & \textbf{0.55} & \textbf{0.6} & \textbf{0.65} & \textbf{0.7} & \textbf{0.75} & \textbf{0.8} & \textbf{0.85} & \textbf{0.9} \\ \hline
\multirow{2}{*}{\textbf{Pure Detection}}  & \textbf{Precision} & 0.625    & 0.573    & 0.516    & 0.447    & 0.357    & 0.256    & 0.168    & 0.082    & 0.026   
\\  & \textbf{Recall} & 0.423    & 0.388    & 0.349    & 0.303    & 0.242    & 0.174    & 0.114    & 0.055    & 0.017
\\ \hline
\multirow{2}{*}{\textbf{Tracked w/ SORT}} & \textbf{Precision} & 0.577    & 0.533    & 0.478    & 0.409    & 0.328    & 0.236    & 0.146    & 0.068    & 0.02 
\\ & \textbf{Recall}    & 0.463    & 0.427    & 0.383    & 0.328    & 0.263    & 0.189    & 0.117    & 0.055    & 0.016   \\ \hline
\end{tabular}
\label{table:change_iou}
\end{table*}

\begin{table*}
\caption{Evaluation of event-based detection vs. YOLOv4: varying confidence thresholds @ IoU threshold = 0.5.}
\centering
\begin{tabular}{ccccccccccccccc}
\multicolumn{2}{l}{\multirow{2}{*}{}}                          & \multicolumn{13}{c}{\textbf{Event-Based Detection: Confidence thresholds @ IoU threshold = 0.5}}      \\
\multicolumn{2}{l}{} &
  \textbf{0.3} &  \textbf{0.35} &  \textbf{0.4} &  \textbf{0.45} &  \textbf{0.5} &  \textbf{0.55} &  \textbf{0.6} &  \textbf{0.65} & \textbf{0.7} &  \textbf{0.75} &  \textbf{0.8} &  \textbf{0.85} &  \textbf{0.9}
\\ \hline
\multirow{2}{*}{\textbf{Pure Detection}}  & \textbf{Precision} & 0.577 & 0.658 & 0.688 & 0.72  & 0.756 & 0.787 & 0.818 & 0.849 & 0.885 & 0.916 & 0.956 & 0.996 & 1.0 
\\ & \textbf{Recall}    & 0.463 & 0.414 & 0.404 & 0.394 & 0.383 & 0.37  & 0.352 & 0.327 & 0.294 & 0.239 & 0.156 & 0.044 & 0.001 \\ \hline
\multirow{2}{*}{\textbf{Tracked w/ SORT}} & \textbf{Precision} & 0.625 & 0.611 & 0.643 & 0.681 & 0.718 & 0.747 & 0.784 & 0.817 & 0.844 & 0.857 & 0.878 & 0.88  & 1.0 
\\  & \textbf{Recall}    & 0.423 & 0.45  & 0.441 & 0.432 & 0.423 & 0.408 & 0.391 & 0.368 & 0.33  & 0.274 & 0.19  & 0.055 & 0.001 \\ \hline
\end{tabular}
\label{table:change_conf}
\end{table*}

\subsection{3D object tracking}

\subsubsection{Quantitative}

Using the proposed detection pipeline and dataset, we evaluate the overall accuracy using the Mean Absolute Error (MAE) and the Root Mean Square Error (RMSE) between the predicted object 3D position in the camera reference frame and the ground truth value from the motion-capture system $\mathbf{T}_W^{M_\bullet}$, $\mathbf{T}_W^{M_S}$, and the calibration $\mathbf{T}_C^{M_S}$.

\begin{table}
\caption{Accuracy analysis of 3D object position estimation.}
\centering
\begin{tabular}{ccccccc}
\multicolumn{1}{l}{} & \multicolumn{1}{l}{} & \multicolumn{2}{c}{\textbf{RGB} detection} & & \multicolumn{2}{c}{\textbf{Event} detection$^*$}       \\
\cline{3-4} \cline{6-7}
\multicolumn{1}{l}{} & \multicolumn{1}{l}{} & \textbf{MAE}  & \textbf{RMSE} &  & \textbf{MAE}  & \textbf{RMSE}
\\ \hline
\multirow{4}{*}{\textbf{\begin{tabular}[c]{@{}c@{}}Filtering \\ only \end{tabular}}} & \textbf{X} & 0.283  & 0.513 & & 0.232 & 0.41\\
     & \textbf{Y}  & 0.765 & 0.79  & & 0.86 & 0.888\\
     & \textbf{Z}  & 0.724 & 1.397 & & 0.929 & 1.613\\
     & \textbf{XZ} & 0.828 & 1.488 & & 0.983 & 1.665\\
     \hline 
     \multirow{4}{*}{\textbf{\begin{tabular}[c]{@{}c@{}}Filtering \\ and \\ CVKF \end{tabular}}} & \textbf{X} & 0.281 & 0.484 & & 0.243 & 0.434\\
     & \textbf{Y}  & 0.763 & 0.788 & & 0.866 & 0.894\\
     & \textbf{Z}  & 0.727 & 1.712 & & 0.98 & 1.663\\
     & \textbf{XZ} & 0.831 & 1.779 & & 1.033 & 1.719\\
     \hline
     \multicolumn{7}{l}{\scriptsize $^*$ The event evaluation uses only one minute of the dataset}
\end{tabular}
\label{table:pos_metrics}
\end{table}

Table~\ref{table:pos_metrics} shows the results from both detection methods.
To achieve an unbiased evaluation of the 3D estimation framework, the event-based prediction uses a manually selected one-minute portion of the ``dynamic.bag" ROSBag where the RVT detector performs well, while the RGB-based pipeline uses the full ROSBag.
The ground truth for people's positions is at head level, and the LiDAR filtering focuses on the hip level.
Metrics are separated by the individual axis and the XZ plane.
The Y-axis (gravity-aligned axis) error of just under a metre reflects head-vs-hip tracking.
Overall, both modalities result in a range of approximately 0.8 to 1~m MAE in the XZ plane, validating the proposed detection/tracking pipeline.
The larger Z-axis error (depth) compared to the X-axis indicates that scan filtering does not fully isolate the object from the foreground and background, as seen in Figure~\ref{fig:bbox_pc} where the spread of 3D point cloud data contained inside a 2D bounding box is broader along the depth axis.

Given that the RMSE weights heavily outlier errors, the difference between MAE and RMSE suggests that a few tracking results are highly inaccurate, while a majority are accurate.
Curiously, the CVKF does not enhance but rather worsens the final estimates.
Thanks to the vision-based SORT tracking, the output of the LiDAR filtering step is already smooth.
Thus, the inherent delay of the final CVKF's estimate with respect to the true state value can only result in lesser accuracy.
Additionally, occlusions and missed associated detections handled by the 2D SORT algorithm lead to the wrong selection of points in the LiDAR scans - leading to high geometric errors.
This correlates with the disparity between MAE and RMSE.

\subsubsection{Qualitative}

\begin{figure}
    \centering
    \def\vdist{0.5cm}
    \def\hdist{-0.1cm}
    \def\legenddist{-0.1cm}
    \def\legendsize{\scriptsize}
    \def\height{2.5cm}
    \begin{tikzpicture}
        \tikzstyle{legend} = [draw=none, minimum height = 0.2em, text width = 6.5em,  minimum width = 6.5em, align =center, execute at begin node=\setlength{\baselineskip}{8pt}]
    
        \node[] (rgbimg) {\includegraphics[clip, trim=5cm 3.5cm 0cm 0cm, height=\height]{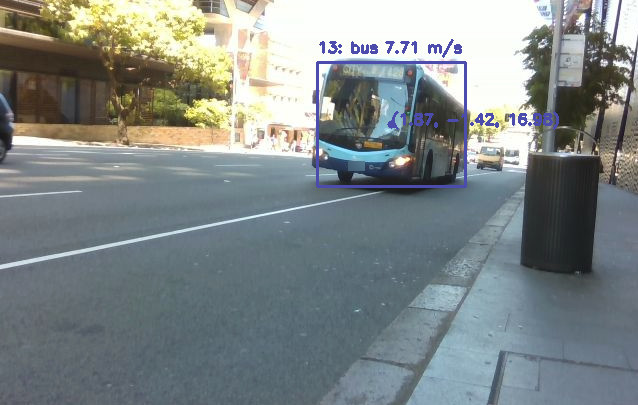}};
        \node[right=\hdist of rgbimg] (rgbbb) {\includegraphics[clip, trim=0cm 0cm 0cm 0cm, height=\height]{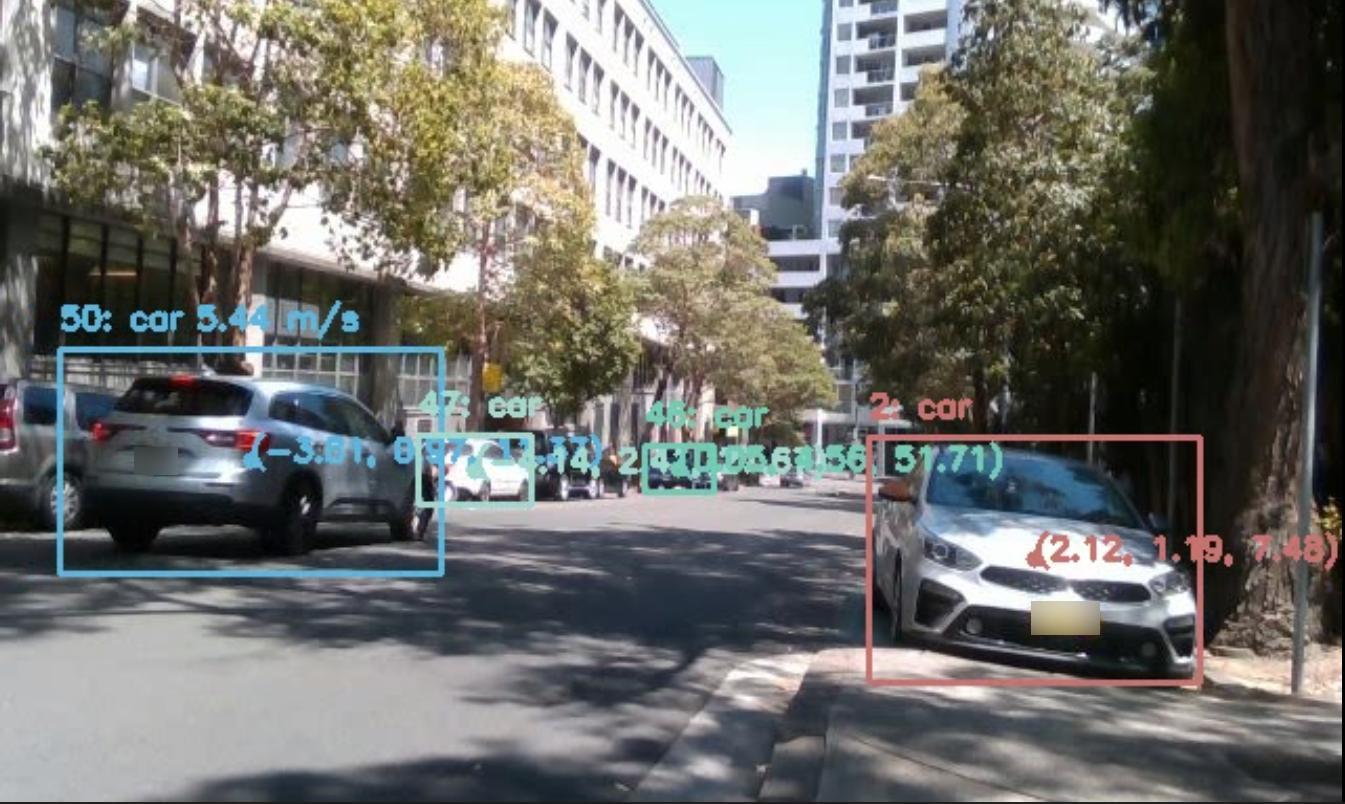}};

        \node[legend, below=\legenddist of rgbimg]{\legendsize (a) Moving bus};
        \node[legend, below=\legenddist of rgbbb]{\legendsize (b) Parked cars};
        
    \end{tikzpicture}
    \vspace{-0.7cm}
    \caption{Detection samples and estimated dynamics in an urban environment.} 
    \label{fig:vlp_estimates}
\end{figure}

To test and demonstrate the ability of the proposed pipeline to provide spatial awareness for assistive robots such as guide-dog-like aid, a sensor suite consisting of an Intel RealSense camera and Velodyne VLP-16 LiDAR was utilised in an urban environment.
This data was collected to inspect the effectiveness of the RGB version of our pipeline on longer-range detections and the ability to estimate vehicle dynamics.
While no ground truth is available for quantitative evaluation, Fig.\ref{fig:vlp_estimates} shows multiple detections of pedestrians and vehicles and their estimated velocity.
These correspond to the expected velocities of the difference agents.

\section{Discussion}
In the LiDAR filtering step, assuming the ``central square" cropping of 3D points aligns the object's centre with the bounding box centre is arbitrary and often untrue.
The central part of the bounding box may correspond to background information.
For instance, when a person extends an arm, the bounding box expands, which shifts the centre away from the torso, leading to an incorrect point selection.
Similarly, for vehicles, the bounding box centre may align with the windshield, causing the LiDAR to observe the background.
Future work will explore using efficient per-pixel semantic labels to better handle partial occlusions.

Our pipeline also assumes constant-velocity models in different trackers.
While effective for wheel-based systems like autonomous vehicles, these models fall short for handheld devices, such as in our dataset, and platforms with jerky motion, such as bipedal and quadrupedal robots.
In scenarios mimicking guide dogs, inaccurate vehicle tracking can have severe consequences.
Investigating varied motion models and incorporating the robot's movement commands might enhance system robustness.
Furthermore, employing trackers like DeepSORT~\cite{Wojke2017-DeepSORT}, which utilise metrics other than IoU, can strengthen frame-to-frame association and tracking.

Expanding into 3D detection improves reliability by merging and cross-checking outputs from multiple modalities, beyond using LiDAR for depth.
However, LiDAR-only detection faces limitations like vertical sparseness and motion distortion.
Sensor fusion is crucial for robust robotic autonomy. Future research should refine detection synchronisation, moving beyond timestamp-based LiDAR scan matching.

Our findings indicate that event-based object detectors lack adaptability and generalisation, while frame-based detectors are ready for use without retraining, thanks to large, diverse publicly available training datasets. The lack of diverse event camera training data hinders adaptability, as event cameras are more affected by camera motion.
Our dataset will enable the robotics community to investigate these issues, paving the way for safer and more robust algorithms.

\section{Conclusion}

This paper has presented a dataset for comparing event-based and RGB-based multi-modal 3D object detection and tracking with LiDAR data.
The dataset includes RGB, event, LiDAR, and inertial data, along with human ground-truth positions determined by a motion-capture system, addressing a gap in publicly available datasets for applications such as guide-dog-like assistive robots.
We proposed a pipeline for dynamic object detection and tracking that performs vision-based object detection followed by LiDAR-based 3D position estimation.
Our experiments show that frame-based detection algorithms generalise well to various scenes, while the current state-of-the-art event models are limited to smaller, automotive-oriented scenarios.

Future work will enrich our dataset with data from various mobile platforms (wheeled, bipedal, and quadrupedal).
This is important due to the spatiotemporal nature of the event data: regular movements lead to recurrent patterns in the event stream.
For the proposed pipeline, our efforts will focus on refining 3D localisation and tracking to better adapt to rapid dynamic changes and employing advanced machine learning techniques for more accurate object isolation.
Ultimately, we will integrate the proposed perception framework into an advanced assistive robot to help vision-impaired users navigate challenging environments safely.

\bibliographystyle{IEEEtran}
\bibliography{references.bib}

\end{document}